\documentclass{article}
\usepackage{spconf,amsmath,graphicx}
\usepackage{multirow}
\usepackage{nicefrac}
\usepackage{scrextend}
\usepackage{booktabs}
\usepackage{amsfonts}

\usepackage{subfiles}

\title{Abnormal Event Detection in Videos using Generative Adversarial Nets}
\name{{\fontsize{11}{50}\selectfont Mahdyar Ravanbakhsh $^{1}$,  Moin Nabi $^{2}$, Enver Sangineto $^{2}$, Lucio Marcenaro$^{1}$, Carlo Regazzoni \sthanks{\scriptsize{Carlo Regazzoni has contributed to produce this work partially under the program ``UC3M-Santander Chairs of Excellence''.}}$^{1,3}$, Nicu Sebe $^{2}$}}
\address{{\fontsize{11.5}{50}\selectfont $^{1}$ DITEN, University of Genova  \hspace{0.7 cm} $^{2}$  DISI, University of Trento \hspace{0.7 cm} $^{3}$ Carlos III University of Madrid}}
\begin{document}
	%
	{\maketitle}
	\begin{abstract}
		In this paper we address the abnormality detection problem in crowded scenes.
		We propose to use Generative Adversarial Nets (GANs), which are trained using {\em normal} frames and corresponding optical-flow images in order to learn an internal representation of the scene {\em normality}. Since our GANs are trained with only normal data, they are not able to generate abnormal events. At testing time the real data are compared with both the appearance and the motion representations reconstructed by our GANs and abnormal areas are detected by computing local differences. Experimental results on challenging abnormality detection datasets show the superiority of the proposed method compared to the state of the art in both frame-level and pixel-level abnormality detection tasks.
	\end{abstract}

	\section{Introduction}
	\label{sec:intro}
	Abnormality detection in crowds is motivated by the increasing interest in video-surveillance systems for public safety. However, despite a lot of research has been done in this area in the past years \cite{li2014anomaly,kim2009observe,Mahadevan.anomaly.2010,mehran2009abnormal,lu2013abnormal,saligrama2012video,cong2011sparse}, the problem is still open.
	
	There are two main reasons for which abnormality detection is challenging. First, existing datasets with {\em ground truth} abnormality samples are small. This limitation is particularly significant for deep-learning based methods, which have shown an impressive accuracy boost in many other recognition tasks \cite{alexnet,DBLP:conf/iccv/Girshick15,donahue2013decaf,Razavian_2014_CVPR_Workshops,DBLP:conf/nips/ZhouLXTO14,DBLP:conf/nips/SimonyanZ14} but are data-hungry. The second reason is the lack of a clear and objective definition of abnormality. Moreover, these two problems are related to each other, because the abnormality definition subjectivity makes it harder to collect abnormality ground truth. 
	
	In order to deal with these problems, {\em generative} methods for abnormality detection focus on modeling only the {\em normal} pattern of the crowd. The advantage of the generative paradigm lies in the fact that only {\em normal} samples are needed at training time, while detection of what is abnormal is based on measuring the distance from the learned normal pattern.
	However, most of the existing generative approaches rely on hand-crafted features to represent visual information \cite{mehran2009abnormal,mousavi2015analyzing,Mahadevan.anomaly.2010,cong2011sparse,kim2009observe} or use Convolutional Neural Networks (CNNs) trained on external datasets \cite{ravanbakhsh2016plug,sabokrouFFK16}. Recently, Xu et al.~\cite{xu2015learning} proposed to use stacked denoising autoencoders. However, the networks used in their work are shallow and based on small image patches. Moreover, additional one-class SVMs need to be trained on top of the learned representation.
	
	In this paper we propose a generative deep learning method applied to abnormality detection in crowd analysis. More specifically, our goal is to use deep networks to learn a representation of the {\em normal pattern} utilizing only {\em normal} training samples, which are much easier to collect. For this purpose, Generative Adversarial Networks (GANs) \cite{DBLP:conf/nips/GoodfellowPMXWOCB14} are used, an emerging approach for training deep networks using only unsupervised data. While GANs are usually used to generate images, we propose to use GANs {\em to learn the normality of the crowd behaviour}.
	At testing time the trained networks are used to generate appearance and motion information. Since our networks have learned to generate {\em only} what is normal, they are not able to reconstruct appearance and motion information of the possible {\em abnormal} regions of the test frame. Exploiting this intuition, a simple difference between the real test-frame representations and the generated descriptions allows us to easily and robustly detect abnormal areas in the frame.
	Extensive experiments on challenging abnormality detection datasets show the superiority of the proposed approach compared to the state of the art.

	\begin{figure}
	\centering
		\begin{minipage}{0.85\linewidth}
			
			\centerline{\includegraphics[width=\linewidth]{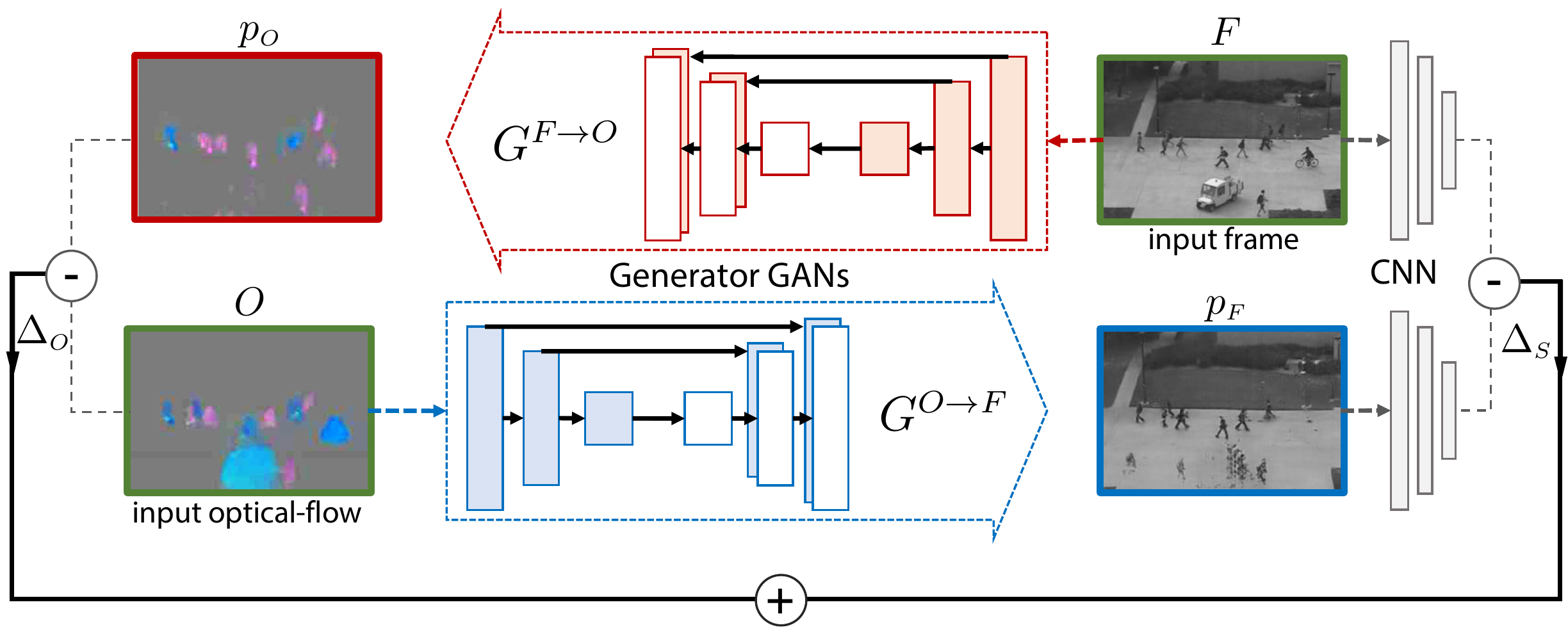}}
		\end{minipage}
		\caption{{\small Top: a generator network takes as input a frame and produces a corresponding optical-flow image. Bottom: a second generator network is fed with a real optical-flow image and outputs an appearance reconstruction.}
		}
		\label{fig:overview}
	\end{figure}
	
	\section{Background}
	\label{sec:format}
	{\bf Abnormality Detection} Our method is different from~\cite{mehran2009abnormal,mousavi2015analyzing,Mahadevan.anomaly.2010,cong2011sparse,kim2009observe,raghavendra2013anomaly,rabiee2016novel,saligrama2012video,lu2013abnormal,mousavi2015abnormality,rabiee2016crowd,rabiee2017detection,huang2016crowd}, which also focus on learning generative models on motion and/or appearance features. A key difference compared to these methods is that they employ hand-crafted features (e.g., Optical-flow, Tracklets, etc.) to model normal-activity patterns, whereas our method learns features from raw-pixels using a deep learning based approach. A deep learning-based approach has been investigated also in~\cite{ravanbakhsh2016plug,sabokrouFFK16}. Nevertheless, these works use existing CNN models trained for other tasks (e.g., object recognition) which are adapted to the abnormality detection task. For instance, Ravanbakhsh et al.~\cite{ravanbakhsh2016plug} propose a Binary Quantization Layer plugged as a final layer on top of a CNN, capturing temporal motion patterns in video frames for the task of abnormality segmentation. Differently from~\cite{ravanbakhsh2016plug}, we specifically propose to train a deep generative network \emph{directly} for the task of abnormality detection. 

Most related to our paper is the work of Xu et al.~\cite{xu2015learning}, who propose to learn motion/appearance feature representations using stacked denoising autoencoders. The networks used in their work are relatively shallow, since training deep autoencoders on small abnormality datasets is prone to over-fitting. Moreover, their networks are not end-to-end trained and the learned representation need externally trained classifiers (multiple one-class SVMs) which are not optimized for the learned features. Conversely, we propose to use adversarial training for our representation learning. Intuitively, the adopted conditional GANs provide data augmentation and implicit data 
 supervision thank to the discriminator network. As a result we can train much deeper generative networks on the same small abnormality datasets and we do not need to train external classifiers.

	\noindent
	{\bf GANs} \cite{DBLP:conf/nips/GoodfellowPMXWOCB14,DBLP:conf/nips/SalimansGZCRCC16,DBLP:journals/corr/RadfordMC15} are based on a two-player game between two different networks, both trained with unsupervised data. One network is the {\em generator} ($G$), which aims at generating realistic data (e.g., images). The second network is the {\em discriminator} ($D$), which aims at discriminating real data from data generated from $G$. 
	Specifically, the {\em conditional} GANs \cite{DBLP:conf/nips/GoodfellowPMXWOCB14}, that we use in our approach,
	take as input an image $x$ and generate a new image $p$. 
	$D$ tries to distinguish $x$ from $p$, while $G$ tries to ''fool'' $D$ producing more and more realistic images which are hard to be distinguished.
	Very recently Isola et al. \cite{DBLP:journals/corr/IsolaZZE16} proposed an ''image-to-image translation'' framework based on conditional GANs, where both $G$ and $D$ are conditioned on the real data. They show that a U-net encoder-decoder with skip connections can be used as the generator architecture together with a patch-based discriminator in order to transform images with respect to different representations.
	A similar framework is adopted here, generating optical-flow images from raw-pixel frames and vice versa. However, we do not aim at generating images which look realistic, but we use $G$ to learn the normal pattern of an observed crowd scene. At testing time, $G$ is used to generate appearance and motion information of the normal content of the input frame.
	Comparing this generated content with the real frame allows us to detect the possible abnormal areas of the frame.
	
	\section{Learning the normal crowd behaviour}
	\label{sec:learing}
	
	We use the framework proposed by Isola et al. \cite{DBLP:journals/corr/IsolaZZE16} to learn the normal behaviour of the observed scene. Specifically, let $F_t$ be the $t$-th frame of a training video and $O_t$ the optical-flow obtained using $F_t$ and $F_{t+1}$. $O_t$ is computed using \cite{brox2004high}.
	We train two networks: ${\cal N}^{F \rightarrow O}$, which generates optical-flow from frames and 
	${\cal N}^{O \rightarrow F}$, which generates frames from optical-flow.
	In both cases, inspired by \cite{DBLP:journals/corr/IsolaZZE16}, our networks are composed of a conditional generator $G$
	and a conditional discriminator $D$ (we refer to \cite{DBLP:journals/corr/IsolaZZE16} for the architectural details of $G$ and $D$). $G$ takes as input an image $x$ and a noise vector $z$ (drawn from a noise distribution ${\cal Z}$) and outputs an image $p = G(x,z)$ of the same dimensions 
	of $x$ 
	but represented in a different channel. 
	For instance, in case of ${\cal N}^{F \rightarrow O}$, $x$ is a frame ($x = F_t$) and $p$ is {\em the reconstruction} of its corresponding optical-flow image $y = O_t$. On the other hand,
	$D$ takes as input two images (either $(x,y)$ or $(x,p)$) and outputs a scalar representing the probability that both its input images came from the real data.

	$G$ and $D$ are trained using both a conditional GAN loss ${\cal L}_{cGAN}$ and a reconstruction loss
	${\cal L}_{L1}$. In case of ${\cal N}^{F \rightarrow O}$, the training set is composed of pairs of frame-optical flow images
	${\cal X} = \{ (F_t, O_t) \}$, where $O_t$ is represented using a standard three-channels representation of the horizontal, the vertical and the magnitude components. 
	${\cal L}_{L1}$ is given by:
	\begin{equation}
	{\cal L}_{L1}(x,y) = ||y - G(x,z) ||_1
	\end{equation}
	\noindent
	while the conditional adversarial loss ${\cal L}_{cGAN}$ is:
	\begin{align}
	{\cal L}_{cGAN}(G,D)= 
	\mathbb{E}_{(x,y) \in {\cal X}} [\log D(x,y)] + \\
	\mathbb{E}_{x \in \{ F_t \}, z \in {\cal Z}} [\log ( 1 - D(x,G(x,z)) )]
	\end{align}
	Conversely, in case of ${\cal N}^{O \rightarrow F}$, we use ${\cal X} = \{ (O_t, F_t) \}$. We refer to \cite{DBLP:journals/corr/IsolaZZE16} for more details about the training procedure. 
	What is important to highlight here is that both $ \{ F_t \}$ and $\{ O_t \}$ are collected 
	using the frames of the only {\em normal} videos of the training dataset. 
	The fact that we do not need videos showing abnormal events at training time makes it possible to train our networks with potentially very large datasets without the need of ground truth samples describing abnormality.
	
	At testing time we use only the generators ($G^{F \rightarrow O}$ and $G^{O \rightarrow F}$) corresponding to the trained networks.
	Since $G^{F \rightarrow O}$ and $G^{O \rightarrow F}$ have observed 
	only normal scenes during training, they are not able to reconstruct an abnormal event. 
	For instance, in Fig.~\ref{fig:overview} (top) a frame $F$, containing a vehicle unusually moving on a University campus,
	is input to $G^{F \rightarrow O}$ and in the generated optical flow image ($p_O$) the abnormal area corresponding to that vehicle is not correctly reconstructed. Similarly, when the real optical flow ($O$) associated with $F$ is input to $G^{O \rightarrow F}$, the network tries to reconstruct the area corresponding to the vehicle but the output is a set of unstructured blobs (Fig.~\ref{fig:overview}, bottom).
	We exploit this {\em inability} of our networks to reliably reconstruct abnormality to detect possible anomalies as explained in the next section.

	\section{abnormality Detection}
	\label{sec:Detection}
	
	At testing time we input $G^{F \rightarrow O}$ and $G^{O \rightarrow F}$ using each frame $F$ of the test video and its corresponding optical-flow image $O$, respectively.
	Note that the random noise vector $z$ is internally produced by the two networks using dropout \cite{DBLP:journals/corr/IsolaZZE16}, and 
	in the following 
	we drop $z$ to simplify our notation.
	Using $F$, an optical-flow reconstruction can be obtained: $p_O= G^{F \rightarrow O}(F)$, which is compared with $O$ using a simple pixel-by-pixel difference, obtaining $\Delta_O = O - p_O$ (see Fig.~\ref{fig:overview}). $\Delta_O$ highlights the (local) differences between the real optical flow and its reconstruction and these differences are higher in correspondence of those areas in which $G^{F \rightarrow O}$ was not able to generate the abnormal behaviour. 
	
	Similarly, we obtain the appearance reconstruction $p_F = G^{O \rightarrow F}(O)$.
	As shown in Fig.~\ref{fig:overview} (bottom), the network generates ''blobs'' in the abnormal areas of 
	$p_F$. Even if these blobs have an appearance completely different from the corresponding area in the real image $F$,
	we empirically observed that a simple pixel-by-pixel difference between $F$ and $p_F$ is less informative than the difference computed in the optical-flow channel. For this reason, a ''semantic'' difference is computed using another network, pre-trained on ImageNet \cite{russakovsky2015imagenet}.
	Specifically, we use AlexNet \cite{alexnet}. Note that AlexNet is trained using supervised data which are pairs of images and object-labels contained in ImageNet. However, no supervision about crowd abnormal behaviour is contained in ImageNet and the network is trained to recognize generic objects.
	Let $h(F)$ be the $conv_5$ representation of $F$ in this network and $h(p_F)$ the corresponding representation of the appearance reconstruction. The fifth convolutional layer of AlexNet (before pooling) is chosen because it represents the input information in a sufficiently abstract space and is the last layer preserving geometric information. 
	We can now compute a semantics-based difference between $F$ and $p_F$: $\Delta_S = h(F) - h(p_F)$.
	
	Finally, $\Delta_S$ and $\Delta_O$ are fused in order to obtain a unique abnormality map. Specifically, we first upsample $\Delta_S$ in order to obtain $\Delta_S'$ with the same resolution as $\Delta_O$. Then, both $\Delta_S'$ and $\Delta_O$ are normalized with respect to their corresponding channel-value range as follows. For each test video $V$
	we compute the maximum value $m_O$ of all the elements of $\Delta_O$ over all the input frames of $V$. The normalized optical-flow difference map is given by:
	\begin{equation}
	N_O(i,j) = 1/m_O \Delta_O(i,j).
	\end{equation}
	\noindent
	Similarly, the normalized semantic difference map $N_S$ is obtained using $m_S$ computed over all the elements of $\Delta_S'$ in all the frames of $V$:
	\begin{equation}
	N_S(i,j) = 1/m_S \Delta_S'(i,j).
	\end{equation}
	The final abnormality map is obtained by summing $N_S$ and $N_O$: $A = N_S + \lambda N_O$. 
	In all our experiments we use $\lambda = 2$. $A$ is our final abnormality heatmap.
	\section{Experimental Results}
	\label{sec:exp}
	\begin{figure}
		
		
		%
		\begin{minipage}[b]{.49\linewidth}
			\centering
			\centerline{\includegraphics[width=\linewidth]{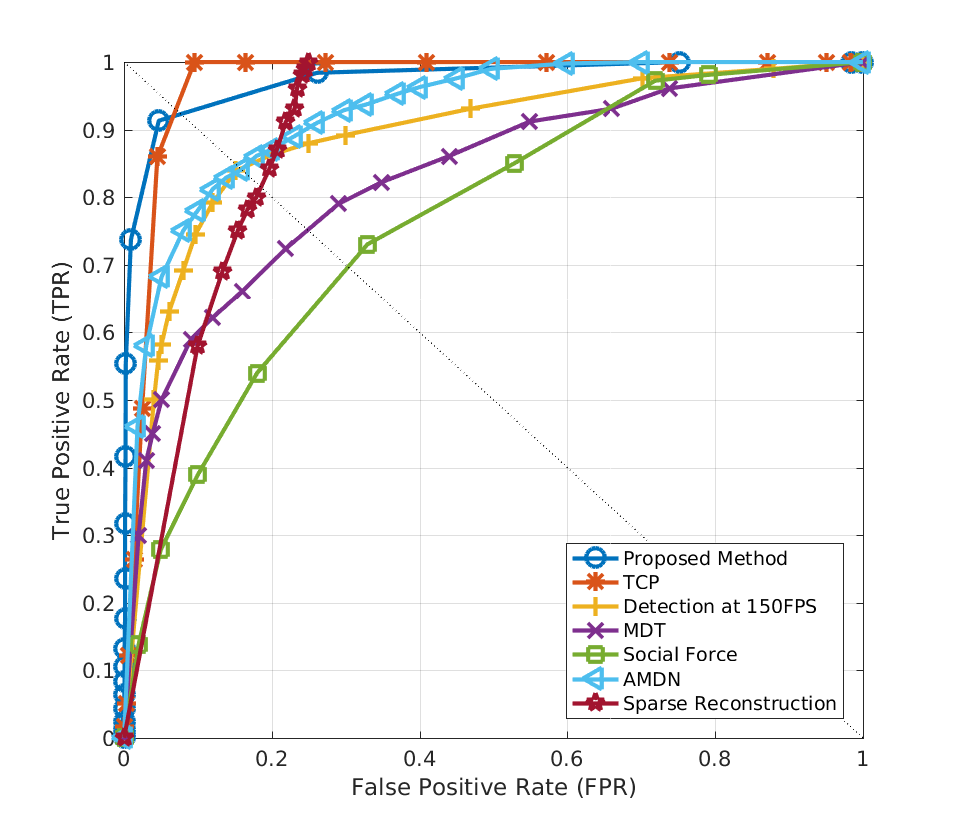}}
			\centerline{\scriptsize(a) Frame-level ROC curves.}\medskip
		\end{minipage}
		\hfill
		\begin{minipage}[b]{0.49\linewidth}
			\centering
			\centerline{\includegraphics[width=\linewidth]{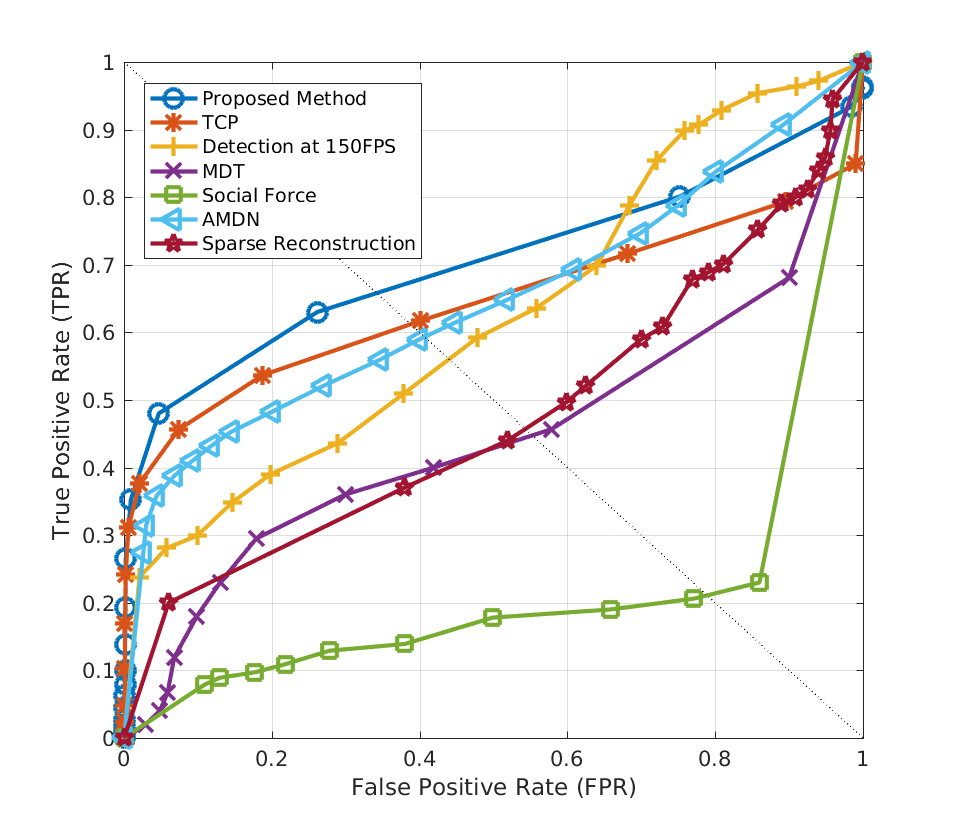}}
			\centerline{\scriptsize(b) Pixel-level ROC curves.}\medskip
		\end{minipage}
		\caption{ROC curves on Ped1 (UCSD dataset).}
		\label{fig:rocfrm}
	\end{figure}
	
	\begin{table*}
		\begin{center}
			\resizebox{0.8\textwidth}{!}{
				\begin{tabular}[width=\textwidth]{l cc l cc l cc}
					\toprule
					\multirow{2}{*}{Method} &\multicolumn{2}{c}{Ped1 (frame-level)} & &\multicolumn{2}{c}{Ped1 (pixel-level)} & &\multicolumn{2}{c}{Ped2 (frame-level)}\\ \cmidrule{2-3} \cmidrule{5-6} \cmidrule{8-9}
					& EER & AUC & & EER & AUC && EER & AUC\\
					\midrule
					MPPCA~\cite{kim2009observe} & 		            40\%& 	59.0\% & 	&	81\%& 20.5\%& 	&	    30\%& 69.3\%\\
					Social force(SF)~\cite{mehran2009abnormal} &    31\% & 	 67.5\% & 	&	79\%& 19.7\%& 	&	    42\%& 55.6\%\\
					SF+MPPCA~\cite{Mahadevan.anomaly.2010} & 		32\% & 	 68.8\%& 	&	71\%& 21.3\%& 	&		36\%& 61.3\%\\
					SR~\cite{cong2011sparse} & 		                19\%&   ---     & 	&	54\%& 45.3\%& 	&	    --- & --- \\
					MDT~\cite{Mahadevan.anomaly.2010} & 		    25\% &  81.8\% & 	&	58\%& 44.1\%& 	&		25\%& 82.9\%\\
					Detection at 150fps~\cite{lu2013abnormal} & 	15\%& 	91.8\% & 	&	43\%& 63.8\%& 	&		--- & ---\\
					Plug-and-Play CNN~\cite{ravanbakhsh2016plug} &  8\% &   95.7\%&     &   40.8\%& 64.5\%& &       18\%& 88.4\%\\
					AMDN (double fusion)~\cite{xu2015learning} & 	16\% & 	 92.1\%& 	&	40.1\%& 67.2\%& &		17\%& 90.8\%\\
					Proposed Method &               \textbf{8\%} & \textbf{97.4\%} &   &\textbf{35\%}& \textbf{70.3\%}& &\textbf{14\%}& \textbf{93.5\%}\\
					\bottomrule
				\end{tabular}
			}
		\end{center}
		\caption{Comparison with the state of the art on the UCSD dataset. The values of the other methods are taken  from~\cite{xu2015learning}.}
		\label{tbl:results}
	\end{table*}
	
	
	\begin{table}
		\begin{center}
			\begin{tabular}[width=0.75\textwidth]{l c}
				\toprule
				Method 										& 	AUC \\
				\midrule
				optical-flow~\cite{mehran2009abnormal}	    &	0.84 \\
				SFM~\cite{mehran2009abnormal} 				&	0.96\\
				Sparse Reconstruction~\cite{cong2011sparse} &   0.97\\
				Commotion~\cite{mousavi2015crowd} 		    &	0.98\\
				Plug-and-Play CNN~\cite{ravanbakhsh2016plug}&	0.98\\
				Proposed Method 							&	\textbf{0.99}\\
				\bottomrule
				
			\end{tabular}
		\end{center}
		\caption{Results on the UMN dataset (all but our values are taken  from~\cite{mousavi2015crowd}).}
		\label{tbl:umn}
	\end{table}
	
	\begin{figure}
		\begin{center}
			\scriptsize{\hspace{0.6cm}real frame  \hspace{0.7cm} generated frame \hspace{0.6cm} generated OF \hspace{0.3cm} abnormality heatmap}
			
			\includegraphics[width=.24\linewidth]{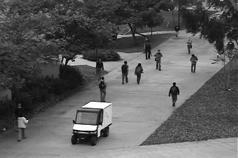}
			\includegraphics[width=.24\linewidth]{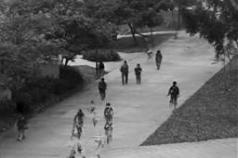}
			\includegraphics[width=.24\linewidth]{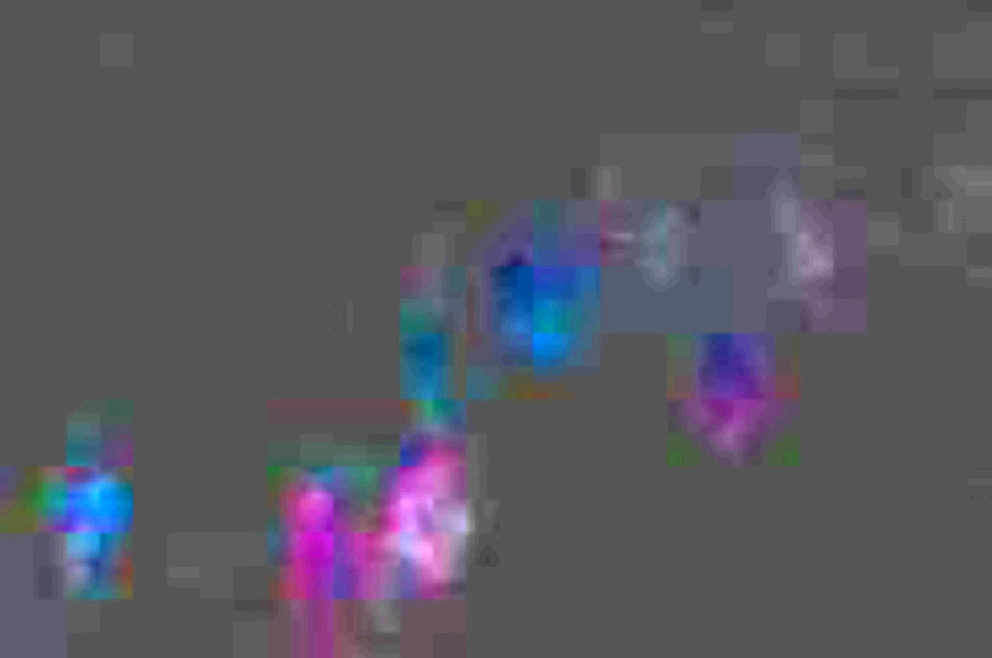}
			\includegraphics[width=.24\linewidth]{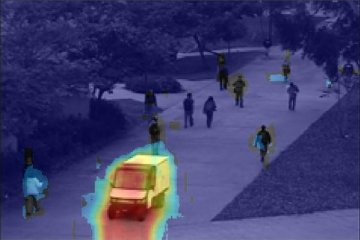}

			\includegraphics[width=.24\linewidth]{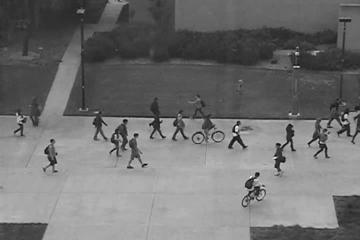}
			\includegraphics[width=.24\linewidth]{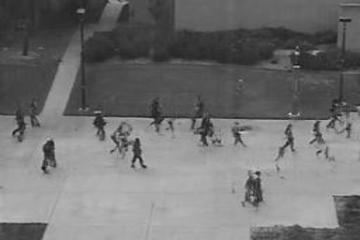}
			\includegraphics[width=.24\linewidth]{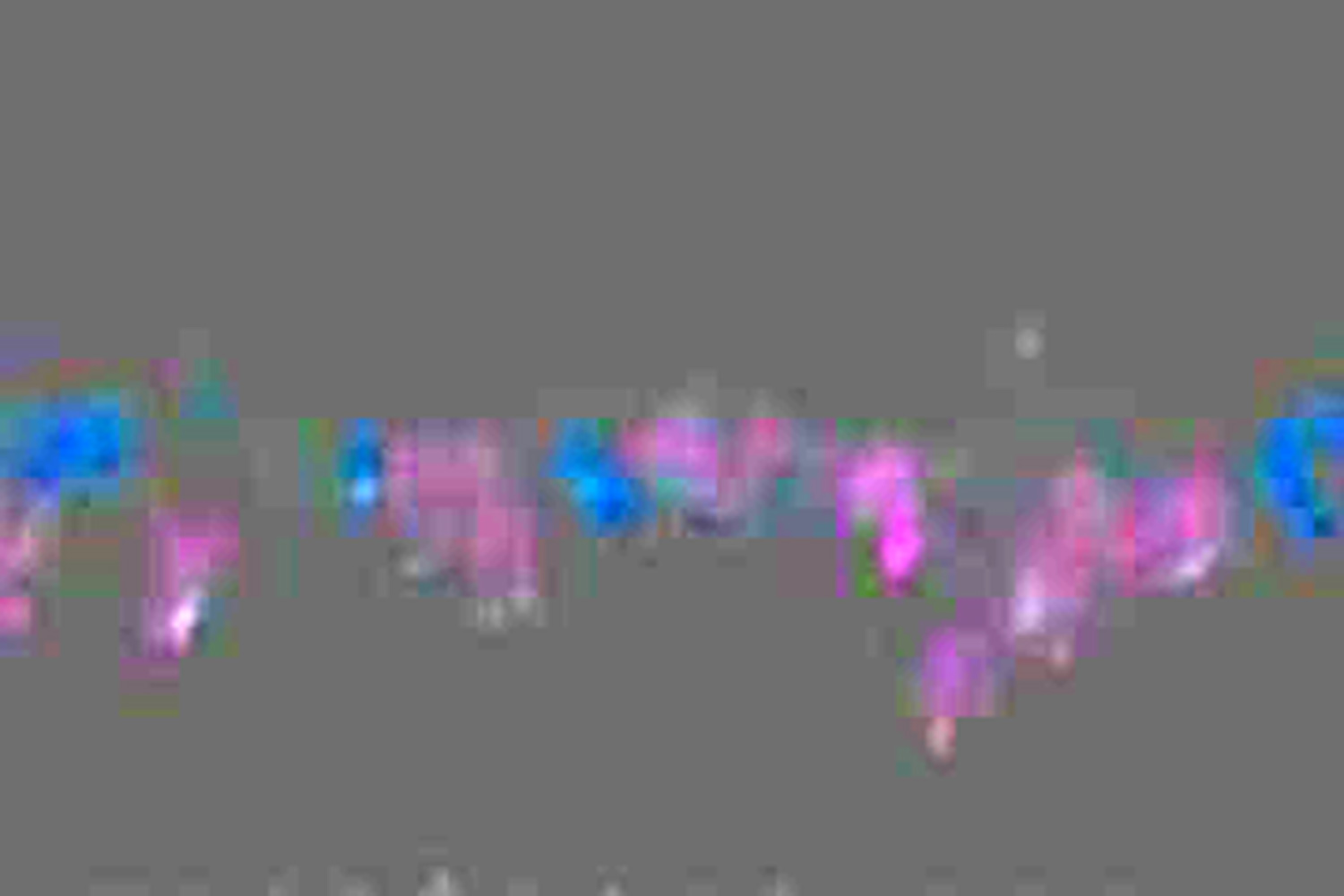}
			\includegraphics[width=.24\linewidth]{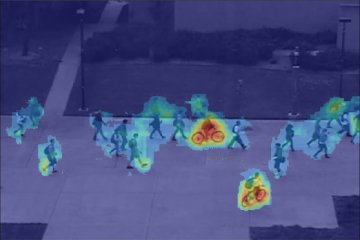}
			\includegraphics[width=.24\linewidth]{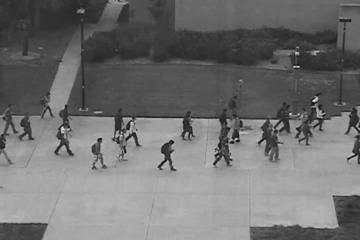}
			\includegraphics[width=.24\linewidth]{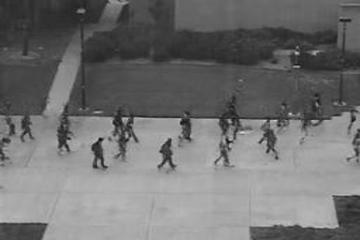}
			\includegraphics[width=.24\linewidth]{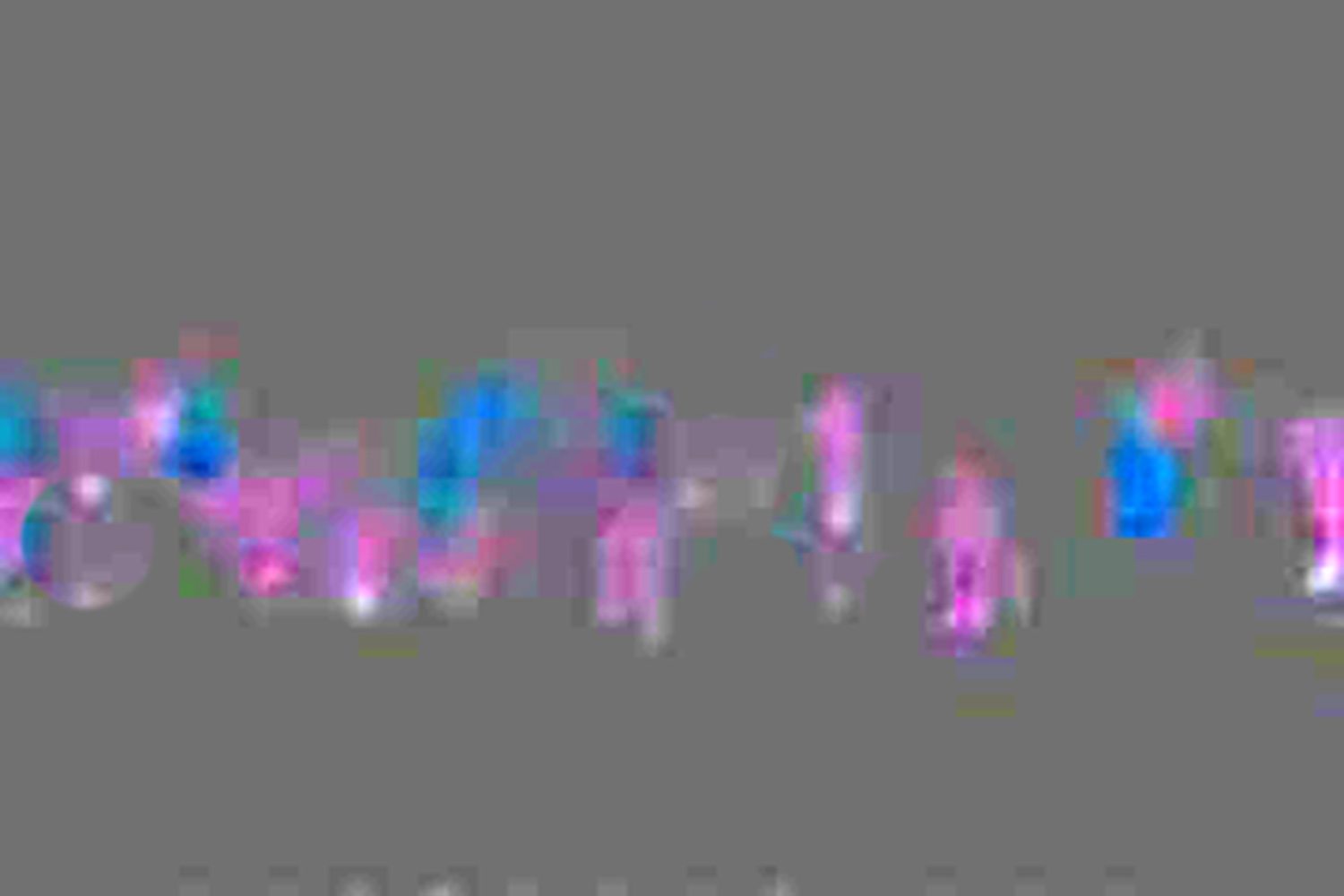}
			\includegraphics[width=.24\linewidth]{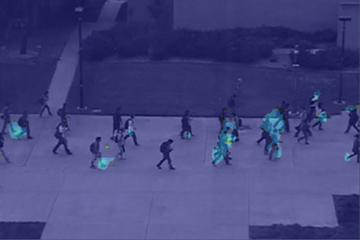}
			
		\end{center}
		\caption{Some examples of abnormality localization on UCSD.}
		\label{fig:pedvis}
	\end{figure}
	
	
	In this section we evaluate our method using two well-known crowd abnormality datasets. We use both a {\em pixel-level} and a {\em frame-level} protocol under the original evaluation setup~\cite{li2014anomaly}. The rest of this section describes the datasets, the experimental setup and the obtained results.\\
	\noindent\textbf{GANs Setup.}
	In our experiments, ${\cal N}^{F \rightarrow O}$ and ${\cal N}^{O \rightarrow F}$ are trained with the train sequences of the UCSD dataset. All frames are resized to $256 \times 256$ pixels. Training is based on stochastic gradient descent with momentum 0.5, batch size 1. Each network is trained for 10 epochs.
	
	\noindent\textbf{Datasets and Experimental Setup.}
	We use two standard datasets: the UCSD Anomaly Detection Dataset~\cite{Mahadevan.anomaly.2010} and the UMN SocialForce~\cite{mehran2009abnormal}. The \textbf{UCSD dataset} is split into two subsets: {\em Ped1}, which contains 34 train and 16 test sequences, and {\em Ped2}, which contains 16 train and 12 test videos. 
	This dataset is challenging due to the low-resolution images, different types of moving objects, the presence of one or more anomalies in the scene. The \textbf{UMN dataset} contains 11 videos in 3 different scenes, with a total amount of 7700 frames. 
	\subsection{Results and Discussion}
	\noindent\textbf{Frame-level abnormality detection.}
	The frame-level abnormality detection criterion is based on checking if the frame contains at least one predicted abnormal pixel: in this case the abnormal label is assigned to the whole frame. The procedure is applied over a range of thresholds to build a ROC curve. We compare our method with the state of the art. Quantitative results using both EER (Equal Error Rate) and AUC (Area Under Curve) are shown in Tab.~\ref{tbl:results}, and the ROC curves in Fig.~\ref{fig:rocfrm}.
	The proposed method is also evaluated on UMN dataset using the same frame level evaluation (Tab.~\ref{tbl:umn}). 
	
	\noindent\textbf{Pixel-level abnormality localization.}
	The goal of the pixel-level evaluation is to measure the accuracy of the abnormality {\em localization}. Following~\cite{li2014anomaly}, 
	a true positive prediction should cover at least 40\% the ground truth abnormal pixels, otherwise the frame is counted as a false positive. 
	Fig.~\ref{fig:rocfrm} shows the ROC curves of the localization accuracy over USDC, and Tab.~\ref{tbl:results} reports a quantitative comparison with the state of the art. The results reported in Tab.~\ref{tbl:results}-\ref{tbl:umn} show that the proposed approach sharply overcomes all the other compared methods.	
	
	\noindent\textbf{Information fusion analysis.}
	In order to analyze the impact on the accuracy provided by each network, ${\cal N}^{O \rightarrow F}$ and ${\cal N}^{F \rightarrow O}$, we perform a set of experiments on UCSD Ped1. In the frame-level evaluation, ${\cal N}^{O \rightarrow F}$ obtains 84.1\% AUC and ${\cal N}^{F \rightarrow O}$ 95.3\% AUC, which are lower than the 97.4\% obtained by the fused version. In the pixel-level evaluation, however, the performance of ${\cal N}^{O \rightarrow F}$ dropped to 30.1\%, while the ${\cal N}^{F \rightarrow O}$ is 66.2\%. 
We believe this is due to the low resolution of $\Delta_S$ (computed over the results obtained using
${\cal N}^{O \rightarrow F}$), which makes the pixel-level localization a hard task.
By fusing appearance and motion we can refine the detected area, which leads to a better localization accuracy. 
	
	\noindent\textbf{Qualitative results.} 
Fig.~\ref{fig:pedvis} shows some results using the standard visualization protocol for abnormality localization (red pixels represent abnormal areas). The figure shows that our approach can successfully localize different abnormality types. Moreover, since the generator learned a spatial distribution of the normal motion in the scene, common perspective issues are automatically alleviated. Fig.~\ref{fig:pedvis} also shows the intuition behind our approach. Normal objects and events (e.g., walking pedestrians) are generated with a sufficient accuracy. However, the generators are not able to reproduce abnormal objects and events (e.g., a vehicle in the first row) and this inability in reproducing abnormalities is what we exploit in order to detect abnormal areas. 

The last row in Fig.~\ref{fig:pedvis} shows a failure case, miss detecting the abnormal object (a skateboard). The failure is probably due to the fact that the skateboard is very small, has a ``normal'' motion (the same speed of normal pedestrians), and is partially occluded.

	\section{Conclusions}
In this paper we addressed the problem of abnormality detection in crowd videos. We proposed a generative deep learning method based on two conditional GANs. 
	Since our GANs are trained using only normal data, they are not able to generate abnormal events. At testing time, a local difference between the real and the generated images is used to detect possible abnormalities.
	Experimental results on standard datasets show that our approach outperforms the state of the art with respect to both the frame-level and the pixel-level evaluation protocols. As future work we will investigate the use of Dynamic Images \cite{Bilen16a} as an alternative to optical-flow in order to represent motion information collected from more than one frame, as suggested by an anonymous reviewer of this paper.

	\bibliographystyle{IEEEbib}
	\bibliography{refs}
	
\end{document}